\documentclass[pdflatex,sn-mathphys-num]{sn-jnl}
\usepackage{graphicx}%
\usepackage{multirow}%
\usepackage{amsmath,amssymb,amsfonts}%
\usepackage{amsthm}%
\usepackage{mathrsfs}%
\usepackage[title]{appendix}%
\usepackage{xcolor}%
\usepackage{textcomp}%
\usepackage{manyfoot}%
\usepackage{booktabs}%
\usepackage{algorithm}%
\usepackage{algorithmicx}%
\usepackage{algpseudocode}%
\usepackage{listings}%
\usepackage{float}

\usepackage{geometry}
\theoremstyle{thmstyleone}%
\theoremstyle{thmstyletwo}%

\theoremstyle{thmstylethree}%

\begin{document}

\title[Article Title]{Analyzing Diversity in Healthcare LLM Research: A Scientometric Perspective}


\author*[1, 2]{\fnm{David} \sur{Restrepo}}\email{davidres@mit.edu}

\author[3]{\fnm{Chenwei} \sur{Wu}}\email{chenweiw@umich.edu}

\author[4, 5]{\fnm{Constanza} \sur{Vásquez-Venegas}}\email{covasquezv@inf.udec.cl}

\author[1]{\fnm{João} \sur{Matos}}\email{jcmatos@mit.edu}

\author[1]{\fnm{Jack} \sur{Gallifant}}\email{jgally@mit.edu}

\author[1, 6, 7]{\fnm{Leo Anthony} \sur{Celi}}\email{lceli@mit.edu}

\author[8]{\fnm{Danielle S.} \sur{Bitterman}}\email{dbitterman@bwh.harvard.edu}

\author[1, 9]{\fnm{Luis Filipe} \sur{Nakayama}}\email{luisnaka@mit.edu}

\affil[1]{\orgdiv{Laboratory for Computational Physiology}, \orgname{Massachusetts Institute of Technology}, \orgaddress{\city{Cambridge}, \state{Massachusetts}, \country{USA}}}

\affil[2]{\orgdiv{Departamento de Telematica}, \orgname{Universidad del Cauca}, \orgaddress{\city{Popayán}, \state{Cauca}, \country{Colombia}}}

\affil[3]{\orgdiv{Department of Electrical Engineering and Computer Science}, \orgname{University of Michigan}, \orgaddress{ \city{Ann Arbor}, \state{Michigan}, \country{USA}}}

\affil[4]{\orgdiv{Scientific Image Analysis Lab, Faculty of Medicine}, \orgname{Universidad de Chile}, \orgaddress{ \city{Santiago}, \state{RM}, \country{Chile}}}

\affil[5]{\orgdiv{Department of Computer Science, Faculty of Engineering}, \orgname{Universidad de Concepción}, \orgaddress{ \city{Concepción}, \state{Biobio}, \country{Chile}}}

\affil[6]{\orgdiv{Department of Biostatistics}, \orgname{Harvard TH Chan School of Public Health}, \orgaddress{ \city{Boston}, \state{Massachusetts}, \country{United States of America}}}

\affil[7]{\orgdiv{Department of Medicine}, \orgname{Beth Israel Deaconess Medical Center}, \orgaddress{ \city{Boston}, \state{Massachusetts}, \country{United States of America}}}

\affil[8]{\orgdiv{Artificial Intelligence in Medicine Program, Mass General Brigham}, \orgname{Harvard Medical School}, \orgaddress{ \city{Boston}, \state{Massachusetts}, \country{USA}}}

\affil[9]{\orgdiv{Department of Ophthalmology}, \orgname{São Paulo Federal University}, \orgaddress{ \city{São Paulo}, \state{São Paulo}, \country{Brazil}}}


\abstract{The deployment of large language models (LLMs) in healthcare has demonstrated substantial potential for enhancing clinical decision-making, administrative efficiency, and patient outcomes. However, the underrepresentation of diverse groups in the development and application of these models can perpetuate biases, leading to inequitable healthcare delivery. This paper presents a comprehensive scientometric analysis of LLM research for healthcare, including data from January 1, 2021, to July 1, 2024. By analyzing metadata from PubMed and Dimensions, including author affiliations, countries, and funding sources, we assess the diversity of contributors to LLM research. Our findings highlight significant gender and geographic disparities, with a predominance of male authors and contributions primarily from high-income countries (HICs). We introduce a novel journal diversity index based on Gini diversity to measure the inclusiveness of scientific publications. Our results underscore the necessity for greater representation in order to ensure the equitable application of LLMs in healthcare. We propose actionable strategies to enhance diversity and inclusivity in artificial intelligence research, with the ultimate goal of fostering a more inclusive and equitable future in healthcare innovation.}




\maketitle

\section{Introduction}\label{sec1}

The advent of large language models (LLMs) represents a significant breakthrough in artificial intelligence (AI), revolutionizing a wide range of applications across numerous fields\cite{liu_summary_2023,chen2023generative, mei2024efficiency, yao2024integrating, xiao2024exploration}. In the context of healthcare, LLMs have demonstrated considerable potential in enhancing clinical decision-making, streamlining administrative processes, and improving patient outcomes through natural language processing tasks such as medical record analysis, automated diagnosis, and personalized treatment recommendations \cite{thirunavukarasu_large_2023}. These advanced models, trained on vast corpora of text data, possess the ability to understand, interpret, and generate human language, thus offering promising solutions to some of the most pressing challenges in modern healthcare \cite{moor_foundation_2023, khurana_natural_2023}.

However, the deployment of LLMs in healthcare is not without its challenges, particularly the need of data and computational resources \cite{fan2024towards}, concerning the issue of representation within the machine learning and healthcare communities \cite{raiaan_review_2024}. Underrepresentation of diverse groups in the development and application of LLMs can lead to the perpetuation or amplification of biases, stereotypes, and negative perceptions of minorities in society \cite{dychiao_large_2024,bender2021dangers,blodgett2020language, smith2022m}. Such biases can adversely affect the quality and equity of healthcare delivery, potentially exacerbating existing disparities \cite{rajkomar_ensuring_2018, restrepo2024seeing}. For instance, if LLMs are predominantly developed and trained by researchers from high-income countries (HICs), the resulting models may not adequately address or understand the healthcare needs of populations in low- and middle-income countries (LMICs). This lack of representation can lead to biased algorithms that fail to generalize across different demographic groups, thus limiting the global applicability and fairness of AI-driven healthcare solutions. Similarly, researchers from HICs are likely to have distinct priorities and blind spots when determining research and developing directions, which may not serve the most pressing needs of diverse populations.

Addressing these challenges requires a concerted effort to promote diversity and inclusion in AI research and development \cite{swartz_science_2019}. One effective approach to highlight and address the lack of representation is through scientometric analysis — a method that involves the quantitative study of science, technology, and innovation \cite{mingers_review_2015}. By analyzing publication patterns, authorship demographics, and funding sources, scientometric analyses can provide critical insights into the evolving landscape of scientific research and its global impact \cite{charpignon2024diversity}. This, in turn, can help identify gaps and biases in the current research landscape and inform strategies to foster greater diversity in the development of LLMs \cite{adebayo_scientometric_2023}.

In this paper, we present a scientometric review of LLM research in healthcare, using data extracted from PubMed and Dimensions API covering the period from January 1, 2021 to July 1, 2024. Through this analysis, we aim to shed light on the diversity of contributors to LLM research in healthcare and underscore the need for broader representation to ensure the equitable and effective application of LLMs worldwide. By evidencing the current state of representation and proposing actionable solutions, our study seeks to contribute to the ongoing discourse on diversity in AI and healthcare, ultimately advocating for a more inclusive and equitable future in health technology innovation.

\section{Methods}\label{sec1}

\subsection{Cohort Selection}

The cohort for this study was defined through a systematic search of PubMed \cite{PubMed} to identify research articles focused on the application of LLMs in healthcare. PubMed was chosen due to its comprehensive coverage of biomedical and life sciences literature, ensuring an extensive collection of relevant studies.

To construct the search query, we used specific terms to capture studies of LLMs and their application in healthcare. The search terms included:

\begin{quote}
    (Large Language Models[Title/Abstract] OR LLMs[Title/Abstract] OR Language Models[Title/Abstract]) AND (Health care[Title/Abstract] OR Healthcare[Title/Abstract] OR Medicine[Title/Abstract] OR Medical[Title/Abstract] OR Clinical[Title/Abstract])
\end{quote}

These terms were designed to be inclusive enough to cover various aspects and terminologies related to LLMs in healthcare. The search was restricted to articles published between January 1, 2021, and July 1, 2024, to capture recent advancements and trends in this rapidly evolving field. The retrieved articles formed the initial dataset for our scientometric analysis. The details of each article, including titles, abstracts, and publication dates, were extracted to facilitate further examination and classification. This initial extraction served as the foundation for subsequent steps in our analysis, where we aimed to gather comprehensive metadata about the authors and their affiliations to assess the diversity of research contributions in this domain.

\subsection{Metadata Extraction and Preprocessing}

Following the initial identification of relevant articles through PubMed, the next step involved the extraction of detailed metadata from the Dimensions database \cite{noauthor_dimensions_nodate}. The PubMed ID (PMID) of each article was used to retrieve information about the publications, including details on the authors, their affiliations, and funding sources. This extraction process utilized the Dimensions API, which allowed for efficient retrieval of metadata by querying the database with the specific PMIDs of the selected articles.

Details about the funding organizations and the countries of the funder organizations were included. To facilitate the analysis, we performed several preprocessing steps:

\begin{itemize}
    \item \textbf{Gender Identification}: The Genderize.io API \cite{Genderize} was employed to infer the gender of the authors based on information such as their first names, last names, and country. This API uses a large database of names to provide probabilistic estimates of gender, allowing us to classify authors as male or female using the maximum probability.
    \item \textbf{Country Classification}: Authors' countries were classified according to the World Bank's 2024 income classifications, which categorize countries into HIC or LMIC \cite{noauthor_world_nodate}. This classification enabled us to assess the representation of different income groups in LLM research.
    \item \textbf{Continent Grouping}: For a comprehensive geographic analysis, countries were grouped into continents. The continents used for this categorization were Africa, Asia, Europe, North America, Oceania, and South America. This categorization facilitated an understanding of the regional distribution of LLM research and the identification of any geographical disparities.
\end{itemize}

The preprocessing of the extracted data entailed the cleaning and formatting of the data to ensure its accuracy and consistency. The authors' names and affiliations were parsed to extract relevant information such as the inference of sex and country from the affiliation country. To standardize the country names, a conversion of country codes to their respective ISO alpha-3 formats was performed. The country formatting conversion was performed through the use of the `pycountry` library. To facilitate the analysis, we also grouped the countries into continents, having: Africa, Asia, Europe, North America, Oceania, and South America.

\subsection{Data Analysis}

In order to archive a global analysis of different variables, we conducted analysis of different variables:

\subsubsection{Gender Distributions} 

The number of male and female authors across the dataset and their position in the author list was  used to calculate overall proportions. The overall proportions of male and female authors are calculated as follows:

Let \( N_{\text{male}} \) and \( N_{\text{female}} \) represent the total counts of male and female authors, respectively, in \( D \). The overall proportions \( P_{\text{male}} \) and \( P_{\text{female}} \) are given by:

\begin{equation}
P_{\text{male}} = \frac{N_{\text{male}}}{N_{\text{male}} + N_{\text{female}}}, \quad P_{\text{female}} = \frac{N_{\text{female}}}{N_{\text{male}} + N_{\text{female}}}
\end{equation}

For specific author positions \( p \) (e.g., first author or last author), the proportions \( P_{\text{male}, p} \) and \( P_{\text{female}, p} \) are similarly calculated:

\begin{equation}
P_{\text{male}, p} = \frac{N_{\text{male}, p}}{N_{\text{male}} + N_{\text{female}}}, \quad P_{\text{female}, p} = \frac{N_{\text{female}, p}}{N_{\text{male}} + N_{\text{female}}}
\end{equation}

Finally, to compare these proportions with the overall distribution, we use a proportions z-test \cite{z-test-prop}.

\subsubsection{Continent and Income Group Distributions per author and funding source}

Author and Funding country affiliations were grouped by continents (Africa, Asia, Europe, North America, Oceania, South America) and classified into HIC and LMIC, The distribution of authors and funding sources by continent and income group was analyzed to highlight geographic disparities in LLM research contributions.

The analysis involved grouping papers based on the continent and income group associated with the authors' affiliations and also with the funding sources. The analysis for authors and income groups was performed independently, however, the calculation method was identical.

The distributions were calculated by dividing the number of papers having authors or funding sources from each continent-income group pair by the total number of papers analyzed. The proportion for each group was calculated as:

\begin{equation}
\label{eq:proportion}
P_{i,j} = \frac{N_{i,j}}{\sum_{i,j} N_{i,j}}
\end{equation}

Where:
\begin{itemize}
    \item \( P_{i,j} \) is the proportion of papers having authors or funding sources for continent \( i \) and income group \( j \).
    \item \( N_{i,j} \) is the count of papers authors or funding sources for continent \( i \) and income group \( j \).
    \item The summation \( \sum_{i,j} N_{i,j} \) represents the total number of papers across all continent-income group pairs.
\end{itemize}

To validate the robustness of our results, we conducted a sensitivity analysis using a bootstrap sampling method \cite{bootstrap}. Let \( D \) be the original dataset, where \( D \in \mathbb{R}^n \) and \( n \) represents the total number of observations. The bootstrap method involved the generation of multiple resampled datasets, each denoted as \( D_k \), where \( k \) ranges from 1 to \( K \). In our case k was set to be 1000.

For each iteration \( k \), the dataset \( D_k \) is created by randomly sampling with replacement from the original dataset \( D \). This process ensures that \( D_k \) retains the same dimensionality as \( D \), but with potentially different distributions due to the resampling. The resampling process can be expressed as:

\begin{equation}
D_k = \{d_1^k, d_2^k, \dots, d_n^k\} \quad \text{where} \quad d_i^k \sim D, \quad i = 1, \dots, n
\end{equation}

Once the resampled dataset \( D_k \) is generated, the proportion of papers with authors or founding sources for each continent and income group \( P_{i,j}^k \) is calculated using the equation \ref{eq:proportion}.

After performing \( K \) iterations, the results are aggregated to obtain the mean proportion \( \hat{P}_{i,j} \) across all bootstrap samples. Additionally, to assess the variability and confidence of the estimates, a 95\% confidence interval is computed for each proportion. The mean proportion and confidence intervals are given by:

\begin{equation}
\hat{P}_{i,j} = \frac{1}{K} \sum_{k=1}^{K} P_{i,j}^k
\end{equation}

\begin{equation}
\text{CI}_{95\%}(P_{i,j}) = \left[\text{Percentile}_{2.5\%}(P_{i,j}^k), \text{Percentile}_{97.5\%}(P_{i,j}^k)\right]
\end{equation}

where \( \text{Percentile}_{x\%}(P_{i,j}^k) \) denotes the \( x \)-th percentile of the bootstrap distribution for the proportion \( P_{i,j}^k \).

\subsubsection{Distribution of Publications per Country:} The distribution of publications per country was analyzed by aggregating the number of publications attributed to authors from each country. A choropleth map was used to illustrate the geographic distribution of research contributions.

\subsubsection{Journal Diversity and Index Testing:} Journal diversity was assessed using an index based on Gini impurity or gini diversity index \cite{gini}, measuring the conditional probabilities of an author being of a certain gender, from a specific continent, and from a particular income group, given that they are an author in a specific journal. The inclusion of income group and continent was to avoid HIC countries skewing the distributions per continents. Then, the diversity index \( D_j \) for a given journal \( j \) is calculated using the equation:

\begin{equation}
D_j = 1 - \sum_{i=1}^{n} \left[ P(G=i \mid J=j) \cdot P(C=i \mid J=j) \cdot P(I=i \mid J=j) \right]^2
\end{equation}

where:
\begin{itemize}
    \item \( P(G=i \mid J=j) \) is the probability of an author being of gender \( i \) given they are an author in journal \( j \).
    \item \( P(C=i \mid J=j) \) is the probability of an author being from country \( i \) given they are an author in journal \( j \).
    \item \( P(I=i \mid J=j) \) is the probability of an author being from income group \( i \) given they are an author in journal \( j \).
\end{itemize}

The index ranges from 0 (less diverse) to 1 (more diverse), with higher values indicating greater diversity. The conditional probabilities were computed using the formula:

\begin{equation}
P(X=i \mid J=j) = \frac{N_{ij}}{N_j}
\end{equation}

where \( N_{ij} \) is the number of authors in category \( i \) for journal \( j \), and \( N_j \) is the total number of authors for journal \( j \).

The index was calculated on journals where more than 50 authors published LLM papers. This metric is a key contribution of our analysis, providing a quantitative measure of diversity in scientific publishing.

\section{Results}\label{sec1}

\subsection{Obtained Cohort}

After running the search query on PubMed, we obtained a total of 1,274 articles.

\subsection{Gender Distributions}

The results showed disparities in gender representation across different author positions (overall, first authors, and last authors). As can be seen in table \ref{table1}a, female authors tend to be less represented in the literature representing only 23.84\% of the authors compared with 76.16\%. The distributions are even more skewed for first and last authors with only 3.06\% and 2.71\% female authors respectively, compared with 11.90\% and 14.04\% respectively. As can be seen in table \ref{table1}b and \ref{table1}c, these distributions were further categorized by HICs authors and LMICs authors finding similar patterns.

\begin{table}[ht!]
\caption{Gender Distribution Among Authors by Region}
\centering
\begin{tabular}{|c|c|c|c|}
\hline
\label{table1}
\textbf{Category} & \textbf{Female (Proportion \%)} & \textbf{Male (Proportion \%)} & \textbf{P-Value} \\
\hline
\multicolumn{4}{|c|}{\textbf{Table 1a: Distribution Over All Authors}} \\
\hline
Overall           & 1963 (23.84\%) & 6271 (76.16\%) & -- \\
First author, n (\%) & 252 (3.06\%)  & 980 (11.90\%)  & $<$0.005 \\
Last author, n (\%)  & 223 (2.71\%)  & 1156 (14.04\%) & $<$0.001 \\
\hline
\multicolumn{4}{|c|}{\textbf{Table 1b: Distribution Over HIC Authors}} \\
\hline
Overall           & 1703 (23.55\%) & 5527 (76.45\%) & -- \\
First author, n (\%) & 216 (2.99\%)  & 835 (11.55\%)  & $<$0.05 \\
Last author, n (\%)  & 185 (2.56\%)  & 1027 (14.20\%) & $<$0.001 \\
\hline
\multicolumn{4}{|c|}{\textbf{Table 1c: Distribution Over LMIC Authors}} \\
\hline
Overall           & 262 (25.27\%) & 775 (74.73\%) & -- \\
First author, n (\%) & 37 (3.57\%)  & 155 (14.95\%) & 0.0342 \\
Last author, n (\%)  & 38 (3.66\%)  & 137 (13.21\%) & 0.2357 \\
\hline
\end{tabular}
\label{table1}
\end{table}

\subsection{Distribution of Papers per Continent for LMICs and HICs}

The analysis of the distribution of research papers across continents and income groups (HICs and LMICs) reveals significant disparities. As shown in Table \ref{table2}, accounts for the majority of contributions from HICs, with 57.07\% of the total papers. Europe follows, contributing 20.17\% from HICs. In contrast, contributions from LMICs are markedly lower, with Asia leading at 11.90\%, driven largely by contributions from China. Africa, Oceania, and South America contribute minimally to the global research output, especially among LMICs.

\begin{figure}[ht!]
\label{fig:map}
\centering
\includegraphics[width=\textwidth]{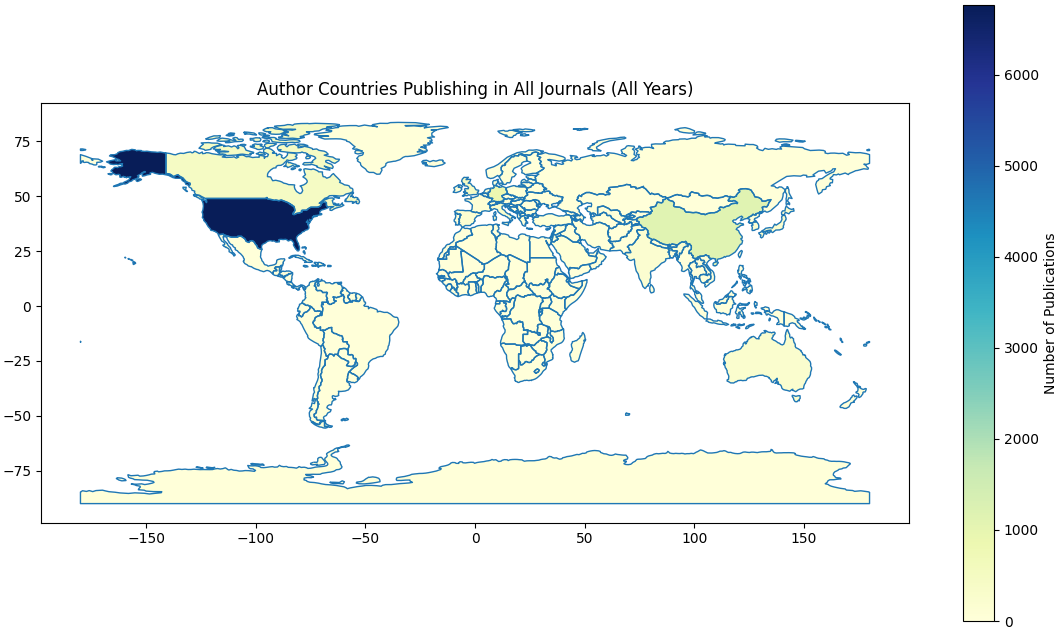}
\caption{World Map Distribution of Research Contributions}
\end{figure}

The world map in Figure \ref{fig:map} provides a geographic visualization of these contributions, further highlighting the dominance of the United States and China in global research, potentially skewing the overall representation of both HICs and LMICs.

Africa's contribution remains minimal, despite representing a significant portion of the world's LMICs, reflecting a broader issue of underrepresentation in global research outputs from this continent.

\begin{table}[ht!]
\centering
\caption{Distribution of Papers by Continent and Income Group}
\label{table2}
\begin{tabular}{|c|c|c|}
\hline
\textbf{Continent} & \textbf{LMIC (Proportion \%)} & \textbf{HIC (Proportion \%)} \\
\hline
Africa & 31 (0.31\%) & 0 (0.00\%) \\
Asia & 1172 (11.90\%) & 774 (7.86\%) \\
Europe & 32 (0.32\%) & 1986 (20.17\%) \\
North America & 0 (0.00\%) & 5620 (57.07\%) \\
Oceania & 0 (0.00\%) & 183 (1.86\%) \\
South America & 40 (0.41\%) & 10 (0.10\%) \\
\hline
\end{tabular}
\end{table}

The bootstrap sensitivity analysis, presented in Table \ref{bootstrap_sensitivity}, further confirms these findings. North America's dominance in the HIC category is robust across different samples, with a mean proportion of 57.11\% (95\% CI: 53.95\%, 60.10\%). Europe's contribution remains substantial as well, with a mean proportion of 20.19\% (95\% CI: 17.56\%, 22.75\%).

On the other hand, the contributions from LMICs, particularly in Africa and South America, are consistently low, even when accounting for sampling variability. The mean proportion for Africa is only 0.33\% (95\% CI: 0.10\%, 0.71\%), and for South America, it is 0.41\% (95\% CI: 0.10\%, 0.91\%). These results reinforce the observation that the distribution of research output is heavily skewed towards certain high-contributing regions, with minimal representation from others.

\begin{table}[ht!]
\centering
\caption{Bootstrap Sensitivity Analysis of Paper Proportions by Continent and Income Group (\%)}
\label{bootstrap_sensitivity}
\begin{tabular}{|c|c|c|}
\hline
\textbf{Continent} & \textbf{LMIC (Mean, CI)} & \textbf{HIC (Mean, CI)} \\
\hline
Africa & 0.33\% (0.10\%, 0.71\%) & 0.00\% (0.00\%, 0.00\%) \\
Asia & 11.91\% (9.91\%, 13.91\%) & 7.80\% (6.10\%, 9.46\%) \\
Europe & 0.34\% (0.10\%, 0.71\%) & 20.19\% (17.56\%, 22.75\%) \\
North America & 0.00\% (0.00\%, 0.00\%) & 57.11\% (53.95\%, 60.10\%) \\
Oceania & 0.00\% (0.00\%, 0.00\%) & 1.85\% (1.11\%, 2.75\%) \\
South America & 0.41\% (0.10\%, 0.91\%) & 0.16\% (0.10\%, 0.40\%) \\
\hline
\end{tabular}
\end{table}

\subsection{Funding Source Analysis}

The raw distribution of funding sources (table \ref{table:funding_distribution}), shows that North America, particularly within the HIC group, dominates the global research funding landscape, contributing 59.15\% of the total funding. Europe follows with 20.42\% of the funding sources, also concentrated in HICs. Asia contributes 11.18\% from LMICs and 7.90\% from HICs, while contributions from other regions such as Africa and South America remain minimal, particularly within the LMIC group.

\begin{table}[ht!]
\centering
\caption{Distribution of Funding Sources by Continent and Income Group}
\label{table:funding_distribution}
\begin{tabular}{|c|c|c|}
\hline
\textbf{Continent} & \textbf{LMIC (Proportion \%)} & \textbf{HIC (Proportion \%)} \\
\hline
Africa & 1 (0.19\%) & 0 (0.00\%) \\
Asia & 58 (11.18\%) & 41 (7.90\%) \\
Europe & 0 (0.00\%) & 106 (20.42\%) \\
North America & 0 (0.00\%) & 307 (59.15\%) \\
Oceania & 0 (0.00\%) & 3 (0.58\%) \\
South America & 2 (0.39\%) & 1 (0.19\%) \\
\hline
\end{tabular}
\end{table}

The bootstrap sensitivity analysis in table \ref{table:bootstrap_funding} confirms North America's dominance in funding is consistently observed, with a mean proportion of 59.73\% (95\% CI: 46.67\%, 72.09\%) for HICs. Europe's contribution remains significant, with a mean proportion of 19.98\% (95\% CI: 9.30\%, 31.38\%) for HICs, reflecting a stable pattern across different resampling scenarios.

In contrast, the contributions from LMICs, particularly in Africa and South America, show greater variability but remain low overall. For instance, Africa's mean proportion of funding from LMICs is 2.00\% (95\% CI: 1.64\%, 3.25\%), while South America's LMIC contribution is 2.12\% (95\% CI: 1.60\%, 3.80\%).

\begin{table}[ht!]
\centering
\caption{Bootstrap Sensitivity Analysis of Funding Source Proportions by Continent and Income Group (\%)}
\label{table:bootstrap_funding}
\begin{tabular}{|c|c|c|}
\hline
\textbf{Continent} & \textbf{LMIC (Mean, CI)} & \textbf{HIC (Mean, CI)} \\
\hline
Africa & 2.00\% (1.64\%, 3.25\%) & 0.00\% (0.00\%, 0.00\%) \\
Asia & 11.11\% (3.70\%, 20.41\%) & 7.92\% (1.87\%, 15.83\%) \\
Europe & 0.00\% (0.00\%, 0.00\%) & 19.98\% (9.30\%, 31.38\%) \\
North America & 0.00\% (0.00\%, 0.00\%) & 59.73\% (46.67\%, 72.09\%) \\
Oceania & 0.00\% (0.00\%, 0.00\%) & 2.18\% (1.61\%, 4.00\%) \\
South America & 2.12\% (1.60\%, 3.80\%) & 2.04\% (1.56\%, 3.74\%) \\
\hline
\end{tabular}
\end{table}


\subsection{Journal Diversity Index}

The diversity of journals assessed using our diversity index based on Gini diversity \cite{gini} calculated for journals with more than 50 authors can be seen in table \ref{tab:index}.

The diversity index for the selected journals reveals significant variation in the representation of different geographic regions and income groups among authors. JMIR Formative Research leads with the highest diversity index of 0.830, indicating a more balanced representation of authors across various regions and income groups. This is closely followed by the Journal of Medical Internet Research and PLOS Digital Health, both of which also show high diversity scores.

On the other end of the spectrum, prestigious journals like Nature and Journal of the American College of Radiology have lower diversity indices, 0.315 and 0.356 respectively, suggesting a more concentrated representation from specific regions or groups, which might indicate a potential bias in the geographical distribution of research contributions.

\begin{table}[ht!]
\caption{Diversity Index for Selected Journals (Ordered from Higher to Lower)}
\label{tab:index}
\centering
\begin{tabular}{|c|c|}
\hline
\textbf{Journal Title} & \textbf{Diversity Index} \\
\hline
\text{JMIR Formative Research} & 0.830 \\
\text{Journal of Medical Internet Research} & 0.825 \\
\text{PLOS Digital Health} & 0.821 \\
\text{Artificial Intelligence in Medicine} & 0.818 \\
\text{JMIR Medical Education} & 0.809 \\
\text{Scientific Reports} & 0.804 \\
\text{Cureus} & 0.785 \\
\text{Journal of Medical Systems} & 0.766 \\
\text{International Journal of Medical Informatics} & 0.760 \\
\text{The Lancet Digital Health} & 0.760 \\
\text{JMIR Medical Informatics} & 0.742 \\
\text{BMC Medical Informatics and Decision Making} & 0.717 \\
\text{JAMA Network Open} & 0.713 \\
\text{Journal of Biomedical Informatics} & 0.685 \\
\text{JAMIA Open} & 0.683 \\
\text{Studies in Health Technology and Informatics} & 0.673 \\
\text{Computers in Biology and Medicine} & 0.645 \\
\text{Nucleic Acids Research} & 0.639 \\
\text{AMIA Annual Symposium Proceedings} & 0.611 \\
\text{Journal of the American Medical Informatics Association} & 0.593 \\
\text{European Archives of Oto-Rhino-Laryngology} & 0.580 \\
\text{bioRxiv} & 0.503 \\
\text{npj Digital Medicine} & 0.502 \\
\text{arXiv} & 0.497 \\
\text{Research Square} & 0.487 \\
\text{Journal of the American Academy of Dermatology} & 0.448 \\
\text{medRxiv} & 0.429 \\
\text{AMIA Joint Summits on Translational Science Proceedings} & 0.412 \\
\text{Journal of Experimental Orthopaedics} & 0.407 \\
\text{Journal of the American College of Radiology} & 0.356 \\
\text{Nature} & 0.315 \\
\hline
\end{tabular}
\end{table}


\section{Discussion}\label{sec1}

The findings of our scientometric analysis reveal disparities in gender and geographic representation in the research on LLMs applications to healthcare. These disparities have important implications for the development and deployment of LLMs in healthcare, particularly with regard to their generalizability, usability, and fairness.

\subsection{Gender Representation}

Our analysis shows that the representation of female authors in LLM research is consistently low across all datasets. Overall, female authors constitute only 23.84\% of the authors, and their representation further diminishes in key author positions, such as first being 3.06\% compared with 11.9\% male authors, and last authors being 2.71\% compared with 14.04\% male authors. This under-representation is observed in both HICs and LMICs. In HICs, female authors make up 23.55\% overall, being 2.99\% first authors, and 2.56\% last authors. In LMICs, the proportions are slightly higher, being 25.27\% overall, 3.57\% first authors, and 3.66\% last authors.

The lower representation of women, particularly in prominent author positions, could limit the diversity of perspectives and expertise in LLM research, potentially leading to biased outcomes in healthcare applications. It is crucial to address these gender disparities to ensure that the development of LLMs benefits from a wide range of viewpoints and experiences, thereby enhancing the models' fairness and applicability across diverse populations.

\subsection{Geographic and Income Group Representation}

The geographic distribution of LLM research contributions highlights a significant imbalance favoring authors from HICs. Our data indicate that North America (HIC) dominates the research landscape, accounting for 57.07\% of the publications, followed by Europe (HIC) with 20.17\%. In contrast, contributions from LMICs are markedly lower. For instance, Asia (LMIC) contributes 11.90\%, and Africa (LMIC) 0.31\%. This skewed distribution raises concerns about the global applicability of LLMs, as models predominantly developed in and for HIC contexts may not address the unique healthcare challenges and needs of LMIC populations.

The disparity is also evident in the funding sources. Unsurprisingly, the majority of research funding comes from organizations based in HICs, with North America (HIC) again leading at 59.15\%, and Europe (HIC) at 20.42\%. LMIC funding contributions are minimal, with Asia (LMIC) at 11.18\%, South America (LMIC) 0.39\% and Africa (LMIC) at just 0.19\%. This funding imbalance can further exacerbate the inequities in research outputs, limiting the ability of LMICs to conduct and contribute to high-quality LLM research. The lack of representation from LMICs, particularly in Latin America, likely arise from the lack of resources for research in the region \cite{restrepo2023scoping}. These challenges represent a significant risk to the generalizability and usability of LLMs in these regions, potentially perpetuating healthcare inequities.

To address these disparities, it is imperative that HICs invest more substantially in funding research in LMICs. Investment from HICs into LMIC research is not only a matter of equity but also of mutual benefit. By funding research in LMICs, HICs can gain access to unique insights and findings that are directly relevant to the global population, including in areas such as computational efficiency, data availability and diversity, and applications such as health practices that may be underrepresented in HIC-centric research.

Moreover, the lack of diverse research inputs from LMICs diminishes the potential impact and innovation that could be achieved if these perspectives were included. Investing in LMIC research will not only strengthen the global health research landscape but will also enhance the generalizability and applicability of research findings across different socio-economic and cultural contexts. In the long run, such investments will contribute to more inclusive and effective LLMs, which can better serve the global community, including populations in HICs that are increasingly diverse and interconnected with global health challenges.

\subsection{Journal Diversity Index}

The novel journal diversity index based on Gini diversity \cite{gini} provides a quantitative measure of diversity within scientific publishing. Our analysis reveals considerable variability in diversity across journals. Journals such as "JMIR Formative Research" (0.830), "Journal of Medical Internet Research" (0.825), and "PLOS Digital Health" (0.821) exhibit high diversity, indicating a more balanced representation of authors from different genders, countries, and income groups. Conversely, journals like "Nature" (0.315) and "Journal of the American College of Radiology" (0.356) show lower diversity, highlighting the need for greater inclusivity.

One possible hypothesis for these disparities in diversity across journals may relate to the internal policies and practices of the journals themselves. Journals with higher diversity indices may have more explicit and proactive policies promoting inclusivity, such as encouraging submissions from underrepresented groups, including authors from LMICs, and adopting editorial practices that prioritize a diverse range of perspectives. These journals might also have more diverse editorial boards, which can influence the types of research that are accepted and the diversity of the authorship. For example, journals like "JMIR Formative Research" and "PLOS Digital Health" may actively work to create an inclusive environment by lowering barriers to entry for authors from diverse backgrounds, including offering fee waivers.

Promoting diversity in scientific journals is essential for democratizing access to healthcare and technologies like LLMs, particularly in low-resource settings. Initiatives aimed at enhancing diversity should be implemented across journals to ensure that research outputs reflect a broad spectrum of perspectives and are applicable to diverse populations worldwide. This can help mitigate biases and enhance the generalizability and fairness of LLMs, ultimately contributing to more equitable healthcare outcomes.

\subsection{Recommendations}

Based on our findings, we recommend several actions to address the disparities in gender and geographic representation in LLM research in healthcare.

\textbf{Institutional policies and initiatives to support women in research:} Academic institutions and research organizations should implement targeted policies and initiatives to encourage and support the participation of women in AI and healthcare research. This can include establishing mentorship programs that pair female researchers with experienced mentors, providing dedicated funding and grants for women-led research projects, and ensuring gender balance in research teams and leadership positions \cite{ceci2024claims}. These initiatives should be accompanied by efforts to address systemic barriers and biases that hinder women's career advancement in research \cite{sato2021leaky}.

\textbf{Equitable distribution of research funds:} Funding bodies and agencies should prioritize the equitable distribution of research funds, supporting projects that involve LMIC researchers and address healthcare challenges pertinent to low-resource settings. This can be achieved through targeted funding calls for LMIC participation, and the inclusion of LMIC researchers in grant review panels. Funding agencies should also consider the unique constraints and challenges faced by LMIC researchers, such as limited access to computational resources and data that are particularly common in LLM research, and provide appropriate support and accommodations\cite{charani2022funders}.

\textbf{Collaborative research networks and partnerships:} To involve researchers from LMICs in global LLM research projects, collaborative research networks and partnerships between HIC and LMIC institutions should be established and strengthened \cite{bowsher2019narrative}. These collaborations can facilitate knowledge transfer, capacity building, and the exchange of expertise and resources. By engaging LMIC researchers as equal partners in research projects, these collaborations can help to ensure that LLM technologies are developed and applied in a manner that is relevant and beneficial to diverse healthcare contexts \cite{meissner2017effect}.

\textbf{Diversity policies in scientific publishing:} Scientific journals and publishing bodies should adopt policies that promote diversity in authorship and editorial boards. This can include implementing diversity metrics, such as tracking the gender and geographic representation of authors and reviewers, and setting targets for improvement. Journals should also encourage submissions from underrepresented regions and groups, potentially through special issues or dedicated submission tracks, such as IEEE ICHI 2024's first workshop of LLM applications in LMIC. Editorial boards should strive for diverse representation and actively seek out expertise from LMICs. Journals, editors, reviewers, and grantors can mandate that author teams disclose their goals for achieving diversity in their research groups and collaborations \cite{jones2022perspective}. This practice, similar to the National Institutes of Health's (NIH) approach for conference grant applications, would promote transparency, accountability, and prioritization of diversity and inclusion in research.

\textbf{Provide formal training on equity and diversity:} Incorporating formal training on equity issues and the importance of diversity in the research process into the syllabi of academic institutions can help educate the next generation of researchers \cite{koolen2017these,zhang2024unified,beddoes2018s}. This training should highlight the benefits of diversity in working groups and published work, such as increased scientific impact and more effective complex problem-solving.

\subsection{Limitations}

While our study provides valuable insights into the diversity of LLM research in healthcare, it is important to acknowledge several limitations. The gender of authors was inferred using the Genderize.io API based on their names' information. This method may introduce biases and inaccuracies, particularly for names that are unisex, culturally specific, or less common. Additionally, it does not account for non-binary and other gender identities. Country information was collected based on authors' affiliations, which may not accurately reflect their nationality or cultural background. This can lead to misclassification, particularly for authors with affiliations in multiple countries or those working abroad.

Our analysis was based on data extracted from PubMed \cite{canese2013pubmed}, which primarily indexes biomedical literature, but also include sources such as ArXiv or MedXiv. Consequently, research articles published in journals not indexed by PubMed or computational conference papers were not included in our study. This exclusion may result in an incomplete representation of the global research landscape.  The study focused on articles published between January 1, 2021, and July 1, 2024. Research trends and representation patterns may evolve over time, and future studies should consider broader temporal scopes to capture these dynamics.

Despite these limitations, our findings highlight critical areas for improvement in promoting diversity and inclusivity in LLM research in healthcare. Addressing these challenges is essential for developing AI technologies that are fair, generalizable, and beneficial to diverse populations worldwide.

\section{Conclusion}

Our scientometric review demonstrates the necessity for greater representation and inclusivity in LLM research in healthcare. Addressing gender and geographic disparities is crucial for developing fair and generalizable AI models that can benefit diverse populations. By promoting diversity in research contributions and funding, and by applying robust measures such us the journal diversity index, we can work towards more equitable and effective healthcare innovations.

In conclusion, this study provides critical insights into the current state of diversity in LLM research and proposes actionable strategies to enhance inclusivity. Future efforts should focus on implementing these strategies to ensure that the benefits of AI-driven healthcare are accessible to all, regardless of gender, geographic location, or economic status.



\end{document}